\documentclass{article} 

\usepackage{microtype}
\usepackage{subfigure}
\usepackage{booktabs} 

\usepackage[colorlinks=true, linkcolor=black, citecolor=black, filecolor=black, urlcolor=black]{hyperref}
\usepackage{enumerate}
\usepackage{url}
\usepackage{amsmath}
\usepackage[draft]{fixme}
\usepackage{graphicx}
\usepackage{xcolor}
\usepackage{tikz}
\usepackage{multicol}
\usepackage[utf8]{inputenc}

\newcommand{\height}{h}
\newcommand{\width}{w}
\newcommand{\modeldim}{d}
\newcommand{\query}{q}

\newcommand{\val}{v}
\newcommand{\key}{k}

\usepackage{ragged2e}
\DeclareRobustCommand{\sidenote}[1]{\marginpar{
                                    \RaggedRight
                                    \textcolor{red}{\textsf{#1}}}}
\usepackage[accepted]{icml2018}

\begin{document}

\twocolumn[
\icmltitle{Image Transformer}



\icmlsetsymbol{equal}{*}

\begin{icmlauthorlist}
\icmlauthor{Niki Parmar *}{g}
\icmlauthor{Ashish Vaswani *}{g}
\icmlauthor{Jakob Uszkoreit}{g}

\icmlauthor{\L{}ukasz Kaiser}{g}
\icmlauthor{Noam Shazeer}{g}
\icmlauthor{Alexander Ku }{b,i}
\icmlauthor{Dustin Tran}{a}
\icmlaffiliation{g}{Google Brain, Mountain View, USA}
\icmlaffiliation{a}{Google AI, Mountain View, USA}
\icmlaffiliation{b}{Department of Electrical Engineering and Computer Sciences, University of California, Berkeley}
\icmlaffiliation{i}{Work done during an internship at Google Brain}
\end{icmlauthorlist}

\icmlcorrespondingauthor{Ashish Vaswani, Niki Parmar, Jakob Uszkoreit}{avaswani@google.com, nikip@google.com, usz@google.com}

\icmlkeywords{Machine Learning, ICML}

\vskip 0.3in
]


\printAffiliationsAndNotice{\icmlEqualContribution} 

\begin{abstract}
Image generation has been successfully cast as an autoregressive sequence generation or transformation problem. Recent work has shown that self-attention is an effective way of modeling textual sequences.
In this work, we generalize a recently proposed model architecture based on self-attention, the Transformer, to a sequence modeling formulation of image generation with a tractable likelihood. By restricting the self-attention mechanism to attend to local neighborhoods we significantly increase the size of images the model can process in practice, despite maintaining significantly larger receptive fields per layer than typical convolutional neural networks. While conceptually simple, our generative models significantly outperform the current state of the art in image generation on ImageNet, improving the best published negative log-likelihood on ImageNet from 3.83 to 3.77.
We also present results on image super-resolution with a large magnification ratio, applying an encoder-decoder configuration of our architecture. In a human evaluation study, we find that images generated by our super-resolution model fool human observers three times more often than the previous state of the art.

\end{abstract}

\vspace{-5mm}
\begin{table}[h!]
\begin{center} 
\begin{tabular}{@{\hspace{.05cm}}c@{\hspace{.05cm}}c@{\hspace{.05cm}}c@{\hspace{.5cm}}c@{\hspace{.05cm}}c@{\hspace{.05cm}}c@{\hspace{.05cm}}c} \\ 
 {\includegraphics[width=.15\linewidth]{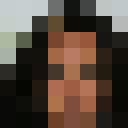}}
& {\includegraphics[width=.15\linewidth]{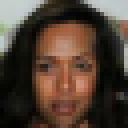}}
& {\includegraphics[width=.15\linewidth]{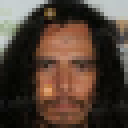}}
& & {\includegraphics[width=.15\linewidth]{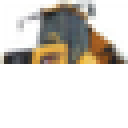}}
& {\includegraphics[width=.15\linewidth]{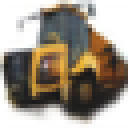}}
& {\includegraphics[width=.15\linewidth]{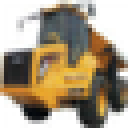}}
\\ [-0.75mm]
{\includegraphics[width=.15\linewidth]{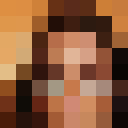}}
& {\includegraphics[width=.15\linewidth]{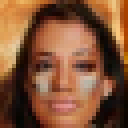}}
& {\includegraphics[width=.15\linewidth]{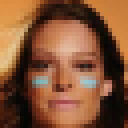}}
& & {\includegraphics[width=.15\linewidth]{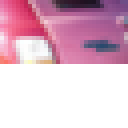}}
& {\includegraphics[width=.15\linewidth]{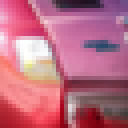}}
& {\includegraphics[width=.15\linewidth]{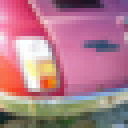}}
\\ [-0.75mm]
 {\includegraphics[width=.15\linewidth]{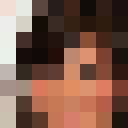}}
& {\includegraphics[width=.15\linewidth]{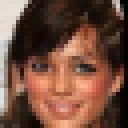}}
& {\includegraphics[width=.15\linewidth]{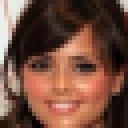}}
& & {\includegraphics[width=.15\linewidth]{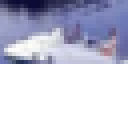}}
& {\includegraphics[width=.15\linewidth]{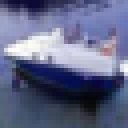}}
& {\includegraphics[width=.15\linewidth]{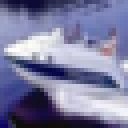}}
\\

\end{tabular}
\caption{Three outputs of a CelebA super-resolution model followed by three image completions by a conditional CIFAR-10 model, with input, model output and the original from left to right}
\end{center}
\end{table}

\section{Introduction}
Recent advances in modeling the distribution of natural images with neural networks allow them to generate increasingly natural-looking images. 
Some models, such as the PixelRNN and PixelCNN \cite{PixelRNN}, have a tractable likelihood. Beyond licensing the comparatively simple and stable training regime of directly maximizing log-likelihood, this enables the straightforward application of these models in problems such as image compression \cite{vandenoord14} and probabilistic planning and exploration \cite{Bellemarre16}.

The likelihood is made tractable by modeling the joint distribution of the pixels in the image as the product of conditional distributions \cite{larochelle2011, Theis2015}. Thus turning the problem into a sequence modeling problem, the state of the art approaches apply recurrent or convolutional neural networks to predict each next pixel given all previously generated pixels \cite{PixelRNN}. Training recurrent neural networks to sequentially predict each pixel of even a small image is computationally very challenging. Thus, parallelizable models that use convolutional neural networks such as the PixelCNN have recently received much more attention, and have now surpassed the PixelRNN in quality \cite{PixelCNN}.

One disadvantage of CNNs compared to RNNs is their typically fairly limited receptive field. This can adversely affect their ability to model long-range phenomena common in images, such as symmetry and occlusion, especially with a small number of layers. Growing the receptive field has been shown to improve quality significantly \cite{PixelCNNpp}. Doing so, however, comes at a significant cost in number of parameters and consequently computational performance and can make training such models more challenging.

In this work we show that self-attention \cite{cheng2016long, decomposableAttnModel, aiayn} can achieve a better balance in the trade-off between the virtually unlimited receptive field of the necessarily sequential PixelRNN and the limited receptive field of the much more parallelizable PixelCNN and its various extensions.

We adopt similar factorizations of the joint pixel distribution as previous work. Following recent work on modeling text \cite{aiayn}, however, we propose eschewing recurrent and convolutional networks in favor of the Image Transformer, a model based entirely on a self-attention mechanism. The specific, locally restricted form of multi-head self-attention we propose can be interpreted as a sparsely parameterized form of gated convolution. By decoupling the size of the receptive field from the number of parameters, this allows us to use significantly larger receptive fields than the PixelCNN.

Despite comparatively low resource requirements for training, the Image Transformer attains a new state of the art in modeling images from the standard ImageNet data set, as measured by log-likelihood. Our experiments indicate that increasing the size of the receptive field plays a significant role in this improvement. We observe significant improvements up to effective receptive field sizes of 256 pixels, while the PixelCNN \cite{PixelCNN} with 5x5 filters used 25.

Many applications of image density models require conditioning on additional information of various kinds: from images in enhancement or reconstruction tasks such as super-resolution, in-painting and denoising to text when synthesizing images from natural language descriptions \cite{Mansimov15}. In visual planning tasks, conditional image generation models could predict future frames of video conditioned on previous frames and taken actions.

In this work we hence also evaluate two different methods of performing conditional image generation with the Image Transformer. In image-class conditional generation we condition on an embedding of one of a small number of image classes. In super-resolution with high magnification ratio (4x), we condition on a very low-resolution image, employing the Image Transformer in an encoder-decoder configuration \cite{KalchbrennerB13}. In comparison to recent work on autoregressive super-resolution \cite{PixelRecursiveSuperResolution}, a human evaluation study found images generated by our models to look convincingly natural significantly more often.








\section{Background}
There is a broad variety of types of image generation models in the literature. This work is strongly inspired by autoregressive models such as fully visible belief networks and NADE \citep{Bengio00, larochelle2011} in that we also factor the joint probability of the image pixels into conditional distributions. Following PixelRNN \citep{PixelRNN}, we also model the color channels of the output pixels as discrete values generated from a multinomial distribution, implemented using a simple softmax layer.

The current state of the art in modeling images on CIFAR-10 data set was achieved by PixelCNN++, which models the output pixel distribution with a discretized logistic mixture likelihood, conditioning on whole pixels instead of color channels and changes to the architecture \citep{PixelCNNpp}. These modifications are readily applicable to our model, which we plan to evaluate in future work.

Another, popular direction of research in image generation is training models with an adversarial loss \citep{gan}. Typically, in this regime a generator network is trained in opposition to a discriminator network trying to determine if a given image is real or generated. In contrast to the often blurry images generated by networks trained with likelihood-based losses, generative adversarial networks (GANs) have been shown to produce sharper images with realistic high-frequency detail in generation and image super-resolution tasks \citep{StackGAN, SRGAN}.

While very promising, GANs have various drawbacks.
They are notoriously unstable \citep{DCGAN}, motivating a large number of methods attempting to make their training more robust \citep{unrolled_gans, began}. Another common issue is that of mode collapse, where generated images fail to reflect the diversity in the training set \citep{unrolled_gans}.

A related problem is that GANs do not have a density in closed-form. This makes it challenging to measure the degree to which the models capture diversity. This also complicates model design. Objectively evaluating and comparing, say, different hyperparameter choices is typically much more difficult in GANs than in models with a tractable likelihood.
\begin{table*}[h]
\begin{center}
\begin{tabular}{@{\hspace{.05cm}}c@{\hspace{.05cm}}c@{\hspace{.05cm}}c@{\hspace{.05cm}}c@{\hspace{.05cm}}c@{\hspace{.3cm}}c@{\hspace{.05cm}}c@{\hspace{.05cm}}c} \\ 
 Input & \multicolumn{3}{c}{Gen} & Truth & Input & Gen & Truth \\
  {\includegraphics[width=.1\linewidth]{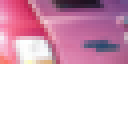}} 
 & {\includegraphics[width=.1\linewidth]{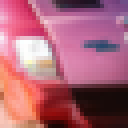}} 
 & {\includegraphics[width=.1\linewidth]{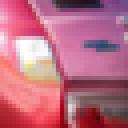}} 
 & {\includegraphics[width=.1\linewidth]{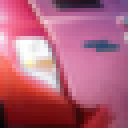}} 
 & {\includegraphics[width=.1\linewidth]{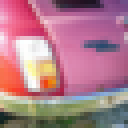}} 
 & {\includegraphics[width=.1\linewidth]{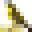}} 
 & {\includegraphics[width=.1\linewidth]{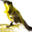}}
 & {\includegraphics[width=.1\linewidth]{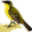}} 
 \\ [-0.75mm]
  {\includegraphics[width=.1\linewidth]{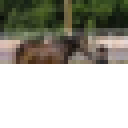}} 
 & {\includegraphics[width=.1\linewidth]{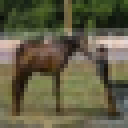}} 
 & {\includegraphics[width=.1\linewidth]{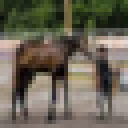}} 
 & {\includegraphics[width=.1\linewidth]{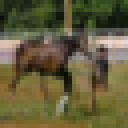}} 
 & {\includegraphics[width=.1\linewidth]{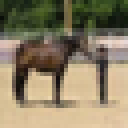}} 
  & {\includegraphics[width=.1\linewidth]{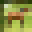}} 
 & {\includegraphics[width=.1\linewidth]{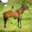}}
 & {\includegraphics[width=.1\linewidth]{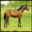}} 
 \\ [-0.75mm]
   {\includegraphics[width=.1\linewidth]{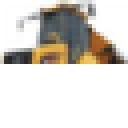}} 
 & {\includegraphics[width=.1\linewidth]{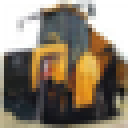}} 
 & {\includegraphics[width=.1\linewidth]{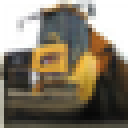}} 
 & {\includegraphics[width=.1\linewidth]{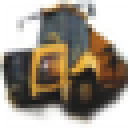}} 
 & {\includegraphics[width=.1\linewidth]{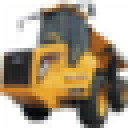}} 
 & {\includegraphics[width=.1\linewidth]{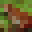}} 
 & {\includegraphics[width=.1\linewidth]{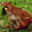}}
 & {\includegraphics[width=.1\linewidth]{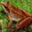}} 
 \\ [-0.75mm]
   {\includegraphics[width=.1\linewidth]{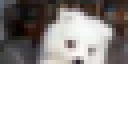}} 
 & {\includegraphics[width=.1\linewidth]{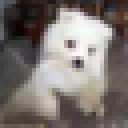}} 
 & {\includegraphics[width=.1\linewidth]{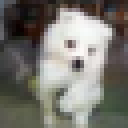}} 
 & {\includegraphics[width=.1\linewidth]{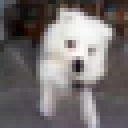}} 
 & {\includegraphics[width=.1\linewidth]{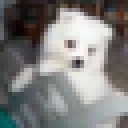}} 
 & {\includegraphics[width=.1\linewidth]{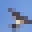}} 
 & {\includegraphics[width=.1\linewidth]{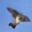}}
 & {\includegraphics[width=.1\linewidth]{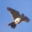}} 
 \\
 \label{tab:completion_and_superres}
\end{tabular} 
\caption{On the left are image completions from our best conditional generation model, where we sample the second half. On the right are samples from our four-fold super-resolution model trained on CIFAR-10. Our images look realistic and plausible, show good diversity among the completion samples and observe the outputs carry surprising details for coarse inputs in super-resolution.}
\end{center}
\end{table*}
\section{Model Architecture}

\subsection{Image Representation}\label{sec:image-rep}
We treat pixel intensities as either discrete categories or ordinal values; this setting depends on the distribution (Section \ref{sub:loss}).
For categories, each of the input pixels' three color channels is encoded using a channel-specific set of $256$ $\modeldim$-dimensional embedding vectors of the intensity values $0-255$. For output intensities, we share a single, separate set of $256$ $\modeldim$-dimensional embeddings across the channels.
For an image of width $w$ and height $h$, we combine the width and channel dimensions yielding a 3-dimensional tensor with shape $[\height, \width \cdot 3, \modeldim]$.

For ordinal values, we run a 1x3 window size, 1x3 strided convolution to combine the $3$ channels per pixel to form an input representation with shape $[\height, \width, \modeldim]$.

To each pixel representation, we add a $\modeldim$-dimensional encoding of coordinates of that pixel. We evaluated two different coordinate encodings: sine and cosine functions of the coordinates, with different frequencies across different dimensions, following~\cite{aiayn}, and learned position embeddings. Since we need to represent two coordinates, we use $\modeldim/2$ of the dimensions to encode the row number and the other $\modeldim/2$ of the dimensions to encode the the column and color channel.

\subsection{Self-Attention}

\begin{figure}
  \centering
  \includegraphics[scale=0.44]{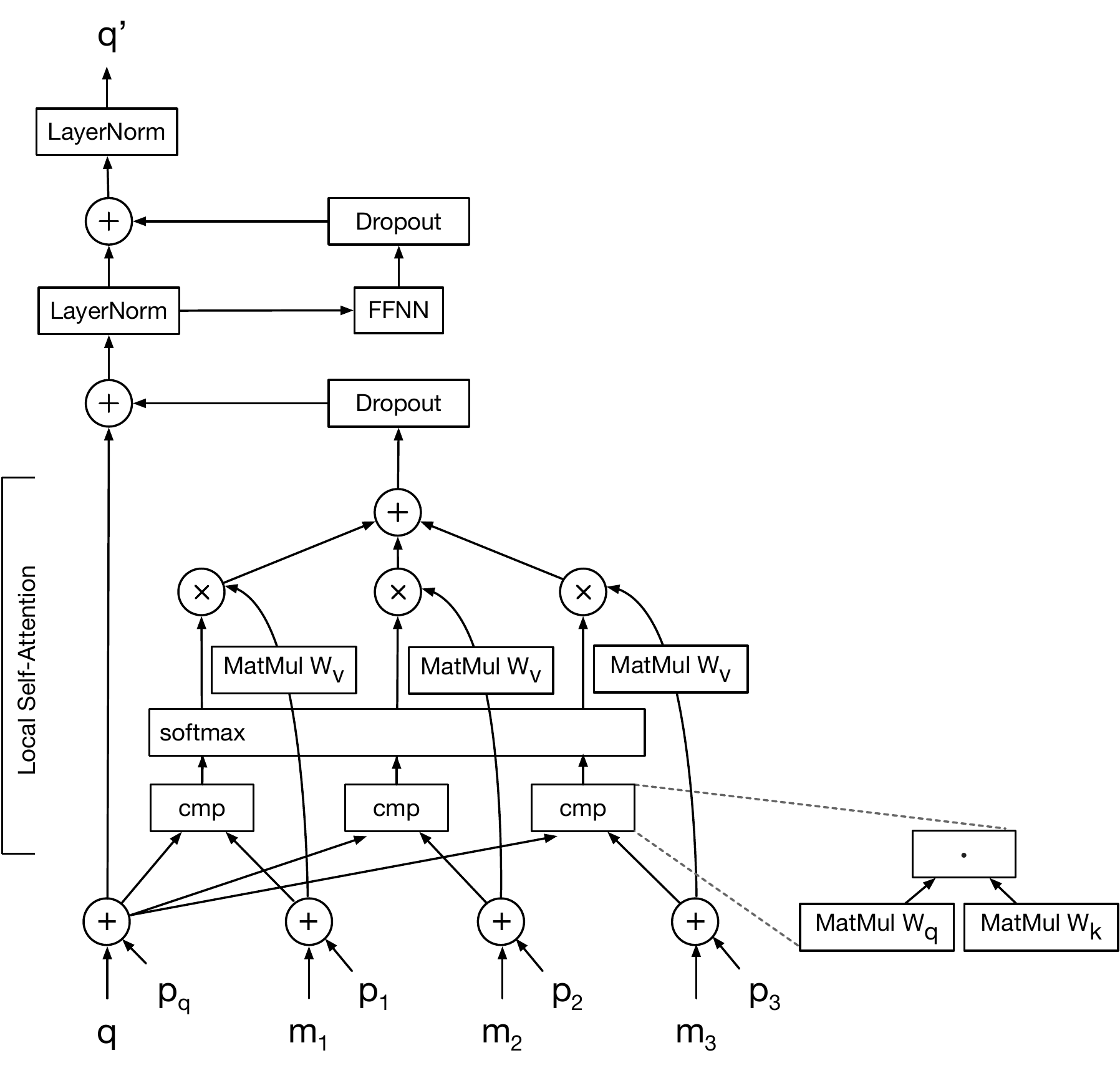}
  \caption{A slice of one layer of the Image Transformer, recomputing the representation $q'$ of a single channel of one pixel $q$ by attending to a memory of previously generated pixels $m_1, m_2, \ldots$. After performing local self-attention we apply a two-layer position-wise feed-forward neural network with the same parameters for all positions in a given layer. Self-attention and the feed-forward networks are followed by dropout and bypassed by a residual connection with subsequent layer normalization. The position encodings $p_q, p_1, \ldots$ are added only in the first layer.}
  \label{fig:model-arch}
\end{figure}

For image-conditioned generation, as in our super-resolution models, we use an encoder-decoder architecture. The encoder generates a contextualized, per-pixel-channel representation of the source image.
The decoder autoregressively generates an output image of pixel intensities, one channel per pixel at each time step. While doing so, it consumes the previously generated pixels and the input image representation generated by the encoder. For both the encoder and decoder, the Image Transformer uses stacks of self-attention and position-wise feed-forward layers, similar to \citep{aiayn}. In addition, the decoder uses an attention mechanism to consume the encoder representation. For unconditional and class-conditional generation, we employ the Image Transformer in a decoder-only configuration.

Before we describe how we scale self-attention to images comprised of many more positions than typically found in sentences, we give a brief description of self-attention.


Each self-attention layer computes a $\modeldim$-dimensional representation for each position, that is, each channel of each pixel. To recompute the representation for a given position, it first compares the position's current representation to other positions' representations, obtaining an attention distribution over the other positions. This distribution is then used to weight the contribution of the other positions' representations to the next representation for the position at hand.

Equations \ref{eqn:self-attention} and \ref{eqn:FFNN} outline the computation in our self-attention and fully-connected feed-forward layers; Figure \ref{fig:model-arch} depicts it. $W_1$ and $W_2$ are the parameters of the feed-forward layer, and are shared across all the positions in a layer.  These fully describe all operations performed in every layer, independently for each position, with the exception of multi-head attention. For details of multi-head self-attention, see \citep{aiayn}.



\begin{multline}
\query_{a} = \mathrm{layernorm}(q + \mathrm{dropout}( \\ 
\mathrm{softmax}\left(\frac{W_{\query} \query (M W_{\key})^T }{\sqrt{d}} \right) M W_{\val}))
\label{eqn:self-attention}
\end{multline}

\begin{equation}
\query' = \mathrm{layernorm}(\query_{a} + \mathrm{dropout} (W_1 \mathrm{ReLu}(W_2 \query_{a})))
\label{eqn:FFNN}
\end{equation}

In more detail, following previous work, we call the current representation of the pixel's channel, or position, to be recomputed the query $q$. The other positions whose representations will be used in computing a new representation for $q$ are $m_1, m_2, \ldots$ which together comprise the columns of the memory matrix $M$.
Note that $M$ can also contain $q$. We first transform $q$ and $M$ linearly by learned matrices $W_q$ and $W_k$, respectively.

The self-attention mechanism then compares $q$ to each of the pixel's channel representations in the memory with a dot-product, scaled by $1/\sqrt{\modeldim}$. We apply the $\mathrm{softmax}$ function to the resulting compatibility scores, treating the obtained vector as attention distribution over the pixel channels in the memory. After applying another linear transformation $W_v$ to the memory $M$, we compute a weighted average of the transformed memory, weighted by the attention distribution. In the decoders of our different models we mask the outputs of the comparisons appropriately so that the model cannot attend to positions in the memory that have not been generated, yet.

To the resulting vector we then apply a single-layer fully-connected feed-forward neural network with rectified linear activation followed by another linear transformation. The learned parameters of these are shared across all positions but different from layer to layer.

As illustrated in Figure\ref{fig:model-arch}, we perform dropout, merge in residual connections and perform layer normalization after each application of self-attention and the position-wise feed-forward networks \citep{layernorm2016, srivastava2014dropout}.

The entire self-attention operation can be implemented using highly optimized matrix multiplication code and executed in parallel for all pixels' channels.




\subsection{Local Self-Attention}
\label{sec:local-self-attention}

The number of positions included in the memory $l_{m}$, or the number of columns of $M$, has tremendous impact on the scalability of the self-attention mechanism, which has a time complexity in $O(h \cdot w \cdot l_{m} \cdot \modeldim)$.

The encoders of our super-resolution models operate on $8\times8$ pixel images and it is computationally feasible to attend to all of their $192$ positions. The decoders in our experiments, however, produce $32\times32$ pixel images with $3072$ positions, rendering attending to all positions impractical.

Inspired by convolutional neural networks we address this by adopting a notion of locality, restricting the positions in the memory matrix $M$ to a local neighborhood around the query position. Changing this neighborhood per query position, however, would prohibit packing most of the computation necessary for self-attention into two matrix multiplications - one for computing the pairwise comparisons and another for generating the weighted averages. To avoid this, we partition the image into query blocks and associate each of these with a larger memory block that also contains the query block. For all queries from a given query block, the model attends to the same memory matrix, comprised of all positions from the memory block.
The self-attention is then computed for all query blocks in parallel.
The feed-forward networks and layer normalizations are computed in parallel for all positions.

\begin{figure}
  \includegraphics[scale=0.265]{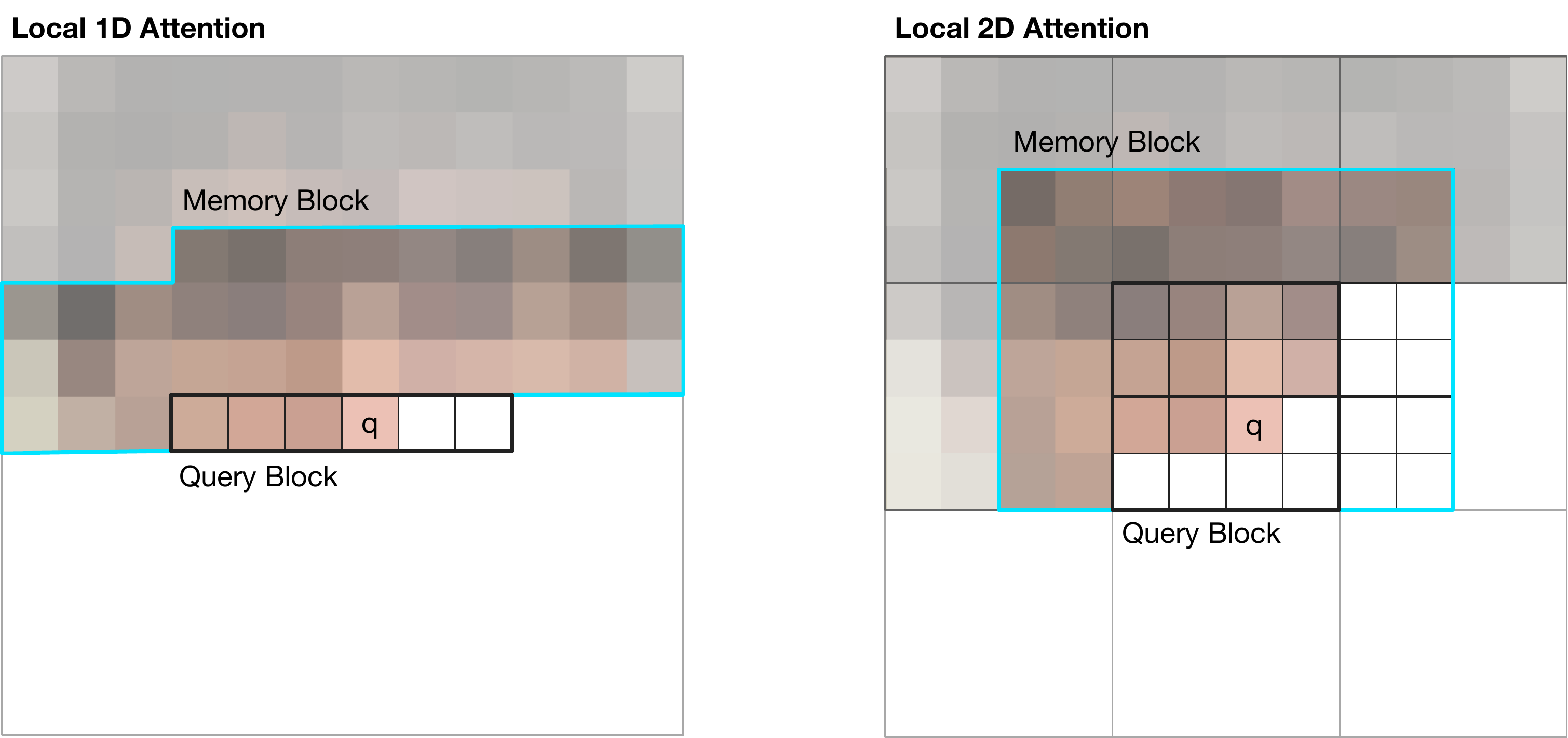}
  \caption{The two different conditional factorizations used in our experiments, with 1D and 2D local attention on the left and right, respectively. In both, the image is partitioned into non-overlapping query blocks, each associated with a memory block covering a superset of the query block pixels.
  In every self-attention layer, each position in a query block attends to all positions in the memory block.
  The pixel marked as $q$ is the last that was generated. All channels of pixels in the memory and query blocks shown in white have masked attention weights and do not contribute to the next representations of positions in the query block. While the effective receptive field size in this figure is the same for both schemes, in 2D attention the memory block
  contains a more evenly balanced number of pixels next to and above the query block, respectively.}
  \label{fig:conditional-factorizations}
\end{figure}

In our experiments we use two different schemes for choosing query blocks and their associated memory block neighborhoods, resulting in two different factorizations of the joint pixel distribution into conditional distributions. Both are illustrated in Figure~\ref{fig:conditional-factorizations}.


\paragraph{1D Local Attention} \label{sec:loc-1d} 
For 1D local attention (Section~\ref{sec:loc-1d}) we first flatten the input tensor with positional encodings in raster-scan order, similar to previous work \citep{PixelRNN}. To compute self-attention on the resulting linearized image, we then partition the length into non-overlapping query blocks $Q$ of length $l_{q}$, padding with zeroes if necessary. While contiguous in the linearized image, these blocks can be discontiguous in image coordinate space. For each query block we build the memory block $M$ from the same positions as $Q$ and an additional $l_{m}$ positions corresponding to pixels that have been generated before, which can result in overlapping memory blocks.

\paragraph{2D Local Attention} \label{sec:loc-2d} In 2D local attention models, we partition the input tensor with positional encodings into rectangular query blocks contiguous in the original image space. We generate the image one query block after another, ordering the blocks in raster-scan order. Within each block, we generate individual positions, or pixel channels, again in raster-scan order.

As illustrated in the right half of Figure~\ref{fig:conditional-factorizations}, we generate the blocks outlined in grey lines left-to-right and top-to-bottom. We use 2-dimensional query blocks of a size $l_q$ specified by height and width $l_q = w_q \cdot h_q$, and memory blocks extending the query block to the top, left and right by $h_m$, $w_m$ and again $w_m$ pixels, respectively.

In both 1D and 2D local attention, we mask attention weights in the query and memory blocks such that positions that have not yet been generated are ignored.

As can be seen in Figure~\ref{fig:conditional-factorizations}, 2D local attention balances horizontal and vertical conditioning context much more evenly. We believe this might have an increasingly positive effect on quality with growing image size as the conditioning information in 1D local attention becomes increasingly dominated by pixels next to a given position as opposed to above it. 



\subsection{Loss Function}
\label{sub:loss}

We perform maximum likelihood, in which we maximize $\log p(x) = \sum_{t=1}^{h\cdot w\cdot 3} \log p(x_t\mid x_{<t})$ with respect to network parameters, and where the
network outputs all parameters of the autoregressive distribution.
We experiment with two settings of the distribution: a categorical distribution across each channel \citep{PixelRNN} and a mixture of discretized logistics over three channels \citep{PixelCNNpp}.

The categorical distribution (\emph{cat}) captures each intensity value as a discrete outcome and factorizes across channels. In total, there are $256\cdot 3=768$ parameters for each pixel; for $32\times 32$ images, the network outputs $786,432$ dimensions.
%

Unlike the categorical distribution, the discretized mixture of logistics (\emph{DMOL}) captures two important properties: the ordinal nature of pixel intensities and simpler dependence across channels \citep{PixelCNNpp}.
For each pixel, the number of parameters is $10$ times the number of mixture components: $10$ for one unnormalized mixture probability, three means,
three standard deviations,
and three coefficients 
which capture the linear dependence.
For 10 mixtures, this translates to $100$ parameters for each pixel; for $32\times 32$ images, the network outputs $102,400$ dimensions, which is a 7x reduction enabling denser gradients and lower memory.

\section{Inference}
Across all of the presented experiments, we use categorical sampling during decoding with a tempered $\mathrm{softmax}$ \citep{PixelRecursiveSuperResolution}. We adjust the concentration of the distribution we sample from with a temperature $\tau > 0$ by which we divide the logits for the channel intensities.

We tuned $\tau$ between $0.8$ and $1.0$, observing the highest perceptual quality in unconditioned and class-conditional image generation with  $\tau=1.0$.
For super-resolution we present results for different temperatures in Table~\ref{tab:CelebASuperResolution}.

\section{Experiments}
All code we used to develop, train, and evaluate our models is available in Tensor2Tensor \citep{tensor2tensor}.

For all experiments we optimize with Adam~\citep{kingma2014adam}, and vary the learning rate as specified in~\cite{aiayn}. We train our models on both p100 and k40 GPUs, with batch sizes ranging from $1$ to $8$ per GPU.

\begin{table}[t]
\centering

\begin{tabular}{@{\hspace{.05cm}}c@{\hspace{.05cm}}c@{\hspace{.05cm}}c@{\hspace{.05cm}}c@{\hspace{.05cm}}c@{\hspace{.3cm}}c@{\hspace{.05cm}}c@{\hspace{.05cm}}c@{\hspace{.05cm}}c@{\hspace{.05cm}}c} \\ 
  {\includegraphics[width=.09\linewidth]{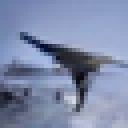}} 
 & {\includegraphics[width=.09\linewidth]{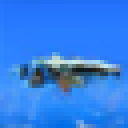}} 
 & {\includegraphics[width=.09\linewidth]{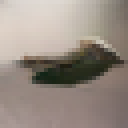}} 
 & {\includegraphics[width=.09\linewidth]{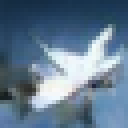}} 
 & {\includegraphics[width=.09\linewidth]{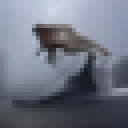}}
&  {\includegraphics[width=.09\linewidth]{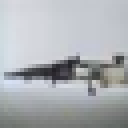}}
& {\includegraphics[width=.09\linewidth]{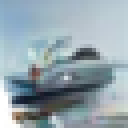}}
& {\includegraphics[width=.09\linewidth]{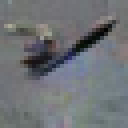}}
& {\includegraphics[width=.09\linewidth]{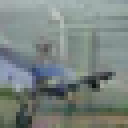}}
& {\includegraphics[width=.09\linewidth]{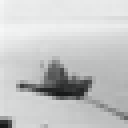}} 
 \\ [-0.75mm]
 {\includegraphics[width=.09\linewidth]{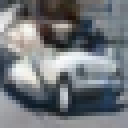}} 
 & {\includegraphics[width=.09\linewidth]{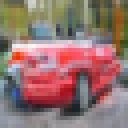}} 
 & {\includegraphics[width=.09\linewidth]{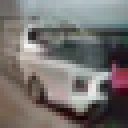}} 
 & {\includegraphics[width=.09\linewidth]{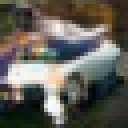}} 
 & {\includegraphics[width=.09\linewidth]{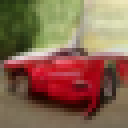}}
 & {\includegraphics[width=.09\linewidth]{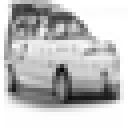}}
& {\includegraphics[width=.09\linewidth]{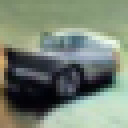}}
& {\includegraphics[width=.09\linewidth]{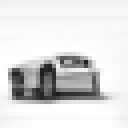}}
& {\includegraphics[width=.09\linewidth]{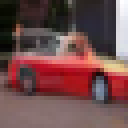}}
& {\includegraphics[width=.09\linewidth]{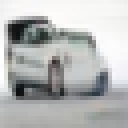}}
\\ [-0.75mm]
{\includegraphics[width=.09\linewidth]{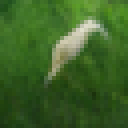}}
& {\includegraphics[width=.09\linewidth]{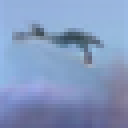}}
& {\includegraphics[width=.09\linewidth]{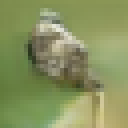}}
& {\includegraphics[width=.09\linewidth]{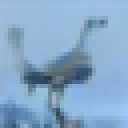}}
& {\includegraphics[width=.09\linewidth]{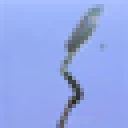}}
& {\includegraphics[width=.09\linewidth]{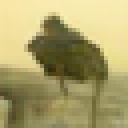}}
& {\includegraphics[width=.09\linewidth]{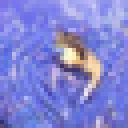}}
& {\includegraphics[width=.09\linewidth]{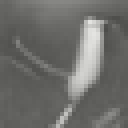}}
& {\includegraphics[width=.09\linewidth]{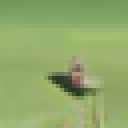}}
& {\includegraphics[width=.09\linewidth]{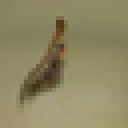}}
\\  [-0.75mm]
{\includegraphics[width=.09\linewidth]{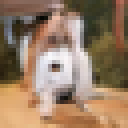}}
& {\includegraphics[width=.09\linewidth]{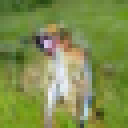}}
& {\includegraphics[width=.09\linewidth]{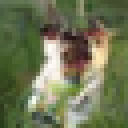}}
& {\includegraphics[width=.09\linewidth]{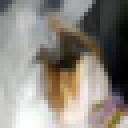}}
& {\includegraphics[width=.09\linewidth]{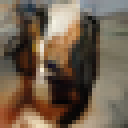}}
&{\includegraphics[width=.09\linewidth]{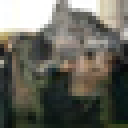}}
& {\includegraphics[width=.09\linewidth]{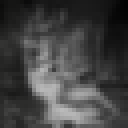}}
& {\includegraphics[width=.09\linewidth]{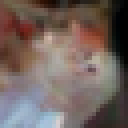}}
& {\includegraphics[width=.09\linewidth]{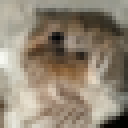}}
& {\includegraphics[width=.09\linewidth]{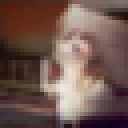}}
\\  [-0.75mm]
{\includegraphics[width=.09\linewidth]{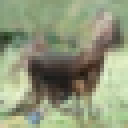}}
& {\includegraphics[width=.09\linewidth]{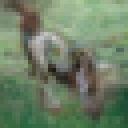}}
& {\includegraphics[width=.09\linewidth]{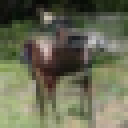}}
& {\includegraphics[width=.09\linewidth]{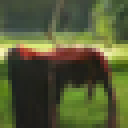}}
& {\includegraphics[width=.09\linewidth]{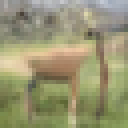}}
& {\includegraphics[width=.09\linewidth]{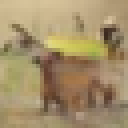}}
& {\includegraphics[width=.09\linewidth]{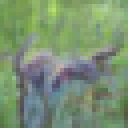}}
& {\includegraphics[width=.09\linewidth]{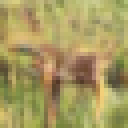}}
& {\includegraphics[width=.09\linewidth]{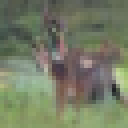}}
& {\includegraphics[width=.09\linewidth]{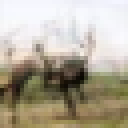}}
\\ [-0.75mm]
{\includegraphics[width=.09\linewidth]{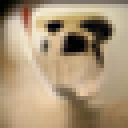}}
& {\includegraphics[width=.09\linewidth]{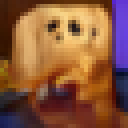}}
& {\includegraphics[width=.09\linewidth]{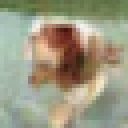}}
& {\includegraphics[width=.09\linewidth]{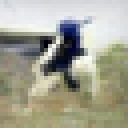}}
& {\includegraphics[width=.09\linewidth]{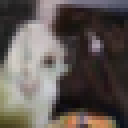}}
& {\includegraphics[width=.09\linewidth]{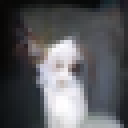}}
& {\includegraphics[width=.09\linewidth]{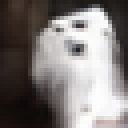}}
& {\includegraphics[width=.09\linewidth]{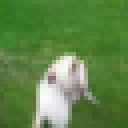}}
& {\includegraphics[width=.09\linewidth]{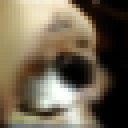}}
& {\includegraphics[width=.09\linewidth]{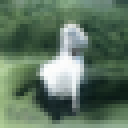}}
\\ [-0.75mm]
{\includegraphics[width=.09\linewidth]{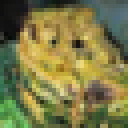}}
& {\includegraphics[width=.09\linewidth]{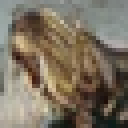}}
& {\includegraphics[width=.09\linewidth]{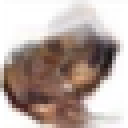}}
& {\includegraphics[width=.09\linewidth]{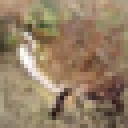}}
& {\includegraphics[width=.09\linewidth]{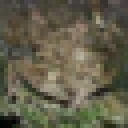}}
& {\includegraphics[width=.09\linewidth]{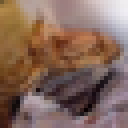}}
& {\includegraphics[width=.09\linewidth]{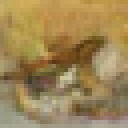}}
& {\includegraphics[width=.09\linewidth]{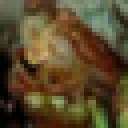}}
& {\includegraphics[width=.09\linewidth]{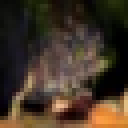}}
& {\includegraphics[width=.09\linewidth]{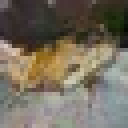}}
\\ [-0.75mm]
{\includegraphics[width=.09\linewidth]{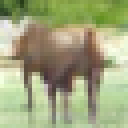}}
& {\includegraphics[width=.09\linewidth]{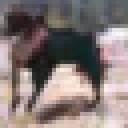}}
& {\includegraphics[width=.09\linewidth]{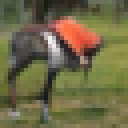}}
& {\includegraphics[width=.09\linewidth]{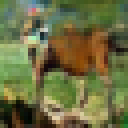}}
& {\includegraphics[width=.09\linewidth]{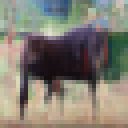}}
& {\includegraphics[width=.09\linewidth]{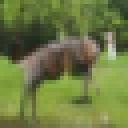}}
& {\includegraphics[width=.09\linewidth]{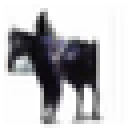}}
& {\includegraphics[width=.09\linewidth]{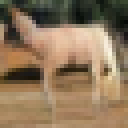}}
& {\includegraphics[width=.09\linewidth]{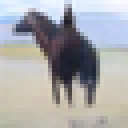}}
& {\includegraphics[width=.09\linewidth]{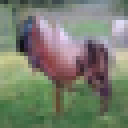}}
\\ [-0.75mm]
{\includegraphics[width=.09\linewidth]{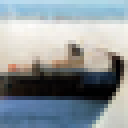}}
& {\includegraphics[width=.09\linewidth]{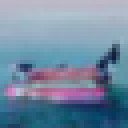}}
& {\includegraphics[width=.09\linewidth]{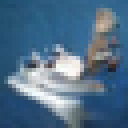}}
& {\includegraphics[width=.09\linewidth]{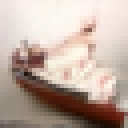}}
& {\includegraphics[width=.09\linewidth]{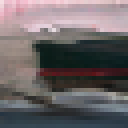}}
& {\includegraphics[width=.09\linewidth]{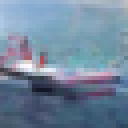}}
& {\includegraphics[width=.09\linewidth]{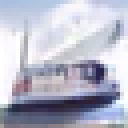}}
& {\includegraphics[width=.09\linewidth]{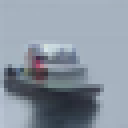}}
& {\includegraphics[width=.09\linewidth]{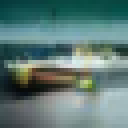}}
& {\includegraphics[width=.09\linewidth]{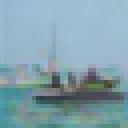}}
\\ [-0.75mm]
 {\includegraphics[width=.09\linewidth]{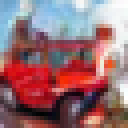}}
& {\includegraphics[width=.09\linewidth]{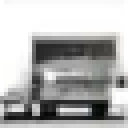}}
& {\includegraphics[width=.09\linewidth]{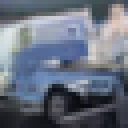}}
& {\includegraphics[width=.09\linewidth]{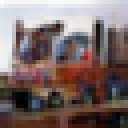}}
& {\includegraphics[width=.09\linewidth]{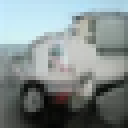}}
& {\includegraphics[width=.09\linewidth]{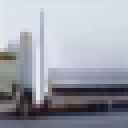}}
& {\includegraphics[width=.09\linewidth]{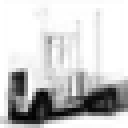}}
& {\includegraphics[width=.09\linewidth]{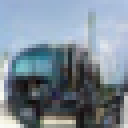}}
& {\includegraphics[width=.09\linewidth]{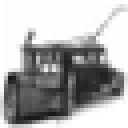}}
& {\includegraphics[width=.09\linewidth]{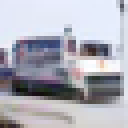}}
\\
\end{tabular} 
\caption{Conditional image generations for all CIFAR-10 categories. Images on the left are from a model that achieves $3.03$ bits/dim on the test set. Images on the right are from our best non-averaged model with $2.99$ bits/dim. Both models are able to generate convincing cars, trucks, and ships. Generated horses, planes, and birds also look reasonable.}
\label{tab:conditional_images}\end{table}

\subsection{Generative Image Modeling}
Our unconditioned and class-conditioned image generation models both use 1D local attention, with $l_q=256$ and a total memory size of $512$.
On CIFAR-10 our best unconditional models achieve a perplexity of $2.90$ bits/dim on the test set using either DMOL or categorical. For categorical, we use $12$ layers with $\modeldim=512$, heads=$4$, feed-forward dimension $2048$ with a dropout of $0.3$. In DMOL, our best config uses $14$ layers, $\modeldim=256$, heads=$8$, feed-forward dimension $512$ and a dropout of $0.2$. This is a considerable improvement over two baselines: the PixelRNN ~\cite{PixelRNN} and PixelCNN++~\cite{PixelCNNpp}. Introduced after the Image Transformer, the also self-attention based PixelSNAIL model reaches a significantly lower perplexity of $2.85$ bits/dim on CIFAR-10 \cite{chen2017pixelsnail}. On the more challenging ImageNet data set, however, the Image Transformer performs significantly better than PixelSNAIL.

We also train smaller $8$ layer CIFAR-10 models which have $\modeldim=512$, $1024$ dimensions in the feed-forward layers, $8$ attention heads and use dropout of $0.1$, and achieve $3.03$ bits/dim, matching the PixelCNN model~\cite{PixelRNN}. Our best CIFAR-10 model with DMOL has $\modeldim$ and feed-forward layer layer dimension of $256$ and perform attention in $512$ dimensions.

ImageNet is a much larger dataset, with many more categories than CIFAR-10, requiring more parameters in a generative model. Our ImageNet unconditioned generation model has $12$ self-attention and feed-forward layers, $\modeldim=512$, $8$ attention heads, $2048$ dimensions in the feed-forward layers, and dropout of $0.1$. It significantly outperforms the Gated PixelCNN and establishes a new state-of-the-art of $3.77$ bits/dim with checkpoint averaging. We trained only unconditional generative models on ImageNet, since class labels were not available in the dataset provided by~\cite{PixelRNN}.

Table~\ref{tab:generative-log-probs} shows that growing the receptive field improves perplexity significantly. We believe this to highlight a key advantage of local self-attention over CNNs: namely that the number of parameters used by local self-attention is independent of the size of the receptive field. Furthermore, while $\modeldim > \mathrm{receptive field}$, self-attention still requires fewer floating-point operations.

For experiments with the categorical distribution we evaluated both coordinate encoding schemes described in Section \ref{sec:local-self-attention} and found no difference in quality. For DMOL we only evaluated learned coordinate embeddings.




%
%

\begin{table}
\centering
\caption{Bits/dim on CIFAR-10 test and ImageNet validation sets. The Image Transformer outperforms all models and matches PixelCNN++, achieving a new state-of-the-art on ImageNet. Increasing memory block size ($bsize$) significantly improves performance.}
\vspace{2mm}

\begin{tabular}{llll}
Model Type & $bsize$ & \multicolumn{2}{c}{NLL}  \\
 & & CIFAR-10 & ImageNet \\
 & & (Test) & (Validation) \\
\hline
Pixel CNN & - & $3.14$ & -\\
Row Pixel RNN & - &  $3.00$ & $3.86$ \\
Gated Pixel CNN & - & $3.03$ & $3.83$\\
Pixel CNN++ & - & $2.92$ & -\\
PixelSNAIL & - & $\mathbf{2.85}$ & $3.80$\\
\hline
Ours 1D local (8l, cat) & 8 & $4.06$ & - \\
 & 16 & $3.47$ & - \\
 & 64 & $3.13$ & - \\
 & 256 & $2.99$ & - \\ 
\hline
Ours 1D local (cat) & 256 & $2.90$ & $\mathbf{3.77}$ \\
Ours 1D local (dmol) & 256 & $2.90$ & -
\label{tab:generative-log-probs}

\end{tabular}
\end{table}

\subsection{Conditioning on Image Class}
We represent the image classes as learned $\modeldim$-dimensional embeddings per class and simply add the respective embedding to the input representation of every input position together with the positional encodings.

We trained the class-conditioned Image Transformer on CIFAR-10, achieving very similar log-likelihoods as in unconditioned generation. The perceptual quality of generated images, however, is significantly higher than that of our unconditioned models. The samples from our $8$-layer class- conditioned models in Table~\ref{tab:conditional_images}, show that we can generate realistic looking images for some categories, such as cars and trucks.



\subsection{Image Super-Resolution}\label{sec:super-res}


Super-resolution is the process of recovering a high resolution image from a low resolution image while generating realistic and plausible details. Following \citep{PixelRecursiveSuperResolution}, in our experimental setup we enlarge an $8 \times 8$ pixel image four-fold to $32 \times 32$, a process that is massively underspecified: the model has to generate aspects such as texture of hair, makeup, skin and sometimes even gender that cannot possibly be recovered from the source image.

Here, we use the Image Transformer in an encoder-decoder configuration, connecting the encoder and decoder through an attention mechanism \citep{aiayn}. For the encoder, we use embeddings for RGB intensities for each pixel in the $8 \times $8 image and add $2$ dimensional positional encodings for each row and width position. Since the input is small, we flatten the whole image as a $[\height \times \width \times 3, \modeldim]$ tensor, where $\modeldim$ is typically $512$. We then feed this sequence to our stack of transformer encoder layers that uses repeated self-attention and feed forward layers. In the encoder we don't require masking, but allow any input pixel to attend to any other pixel. In the decoder, we use a stack of local self-attention, encoder-decoder-attention and feed-forward layers. We found using two to three times fewer encoder than decoder layers to be ideal for this task.

We perform end-to-end training of the encoder-decoder model for Super resolution using the log-likelihood objective function. Our method generates higher resolution images that look plausible and realistic across two datasets.

For both of the following data sets, we resized the image to $8 \times 8$ pixels for the input and $32\times32$ pixels for the label using TensorFlow's $\mathrm{area}$ interpolation method.

\begin{table}[h!]
\caption{Negative log-likelihood and human eval performance for the Image Transformer on CelebA. The fraction of humans fooled is significantly better than the previous state of the art.}
\label{tab:superres_table}
\begin{center}
\vspace{2mm}
\begin{tabular}{lll}
Model Type  & $\tau$ & \%Fooled  \\
\hline
ResNet  & $n/a$ & $4.0$  \\
srez GAN  & $n/a$ & $8.5$  \\
\hline
PixelRecursive & $1.0$  & $11.0$  \\
\citep{PixelRecursiveSuperResolution} & $0.9$  & $10.4$  \\
& $0.8$  & $10.2$ \\
\hline
1D local & $1.0$  & $29.6\pm4.0$ \\
Image Transformer & $0.9$ & $33.5\pm3.5$ \\
& $0.8$ &  $\mathbf{35.94\pm3.0}$ \\
\hline
2D local & $1.0$ & $30.64\pm4$ \\
Image Transformer& $0.9$ & $34\pm3.5$  \\
& $0.8$ &  $\mathbf{36.11\pm2.5}$ \\
\bottomrule
\label{tab:CelebASuperResolution}
\end{tabular}

\end{center}
\end{table}




\begin{table*}[h!]
\label{tab:celeba_images}
\centering
\begin{tabular}{@{\hspace{.05cm}}c@{\hspace{.05cm}}c@{\hspace{.05cm}}c@{\hspace{.05cm}}c@{\hspace{.05cm}}c@{\hspace{.05cm}}c@{\hspace{.05cm}}c@{\hspace{.05cm}}c} \\ 
  Input &  \multicolumn{3}{c} {1D Local Attention} &  \multicolumn{3}{c} {2D Local Attention} & Original \\
  & $\tau=0.8$ & $\tau=0.9$ & $\tau=1.0$ & $\tau=0.8$ & $\tau=0.9$ & $\tau=1.0$ & \\
{\includegraphics[width=.1\linewidth]{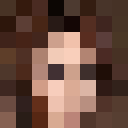}}
& {\includegraphics[width=.1\linewidth]{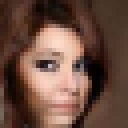}}
& {\includegraphics[width=.1\linewidth]{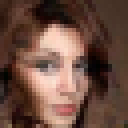}}
& {\includegraphics[width=.1\linewidth]{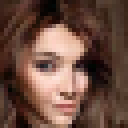}}
& {\includegraphics[width=.1\linewidth]{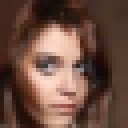}}
& {\includegraphics[width=.1\linewidth]{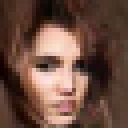}}
& {\includegraphics[width=.1\linewidth]{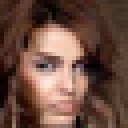}}
& {\includegraphics[width=.1\linewidth]{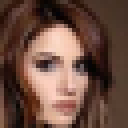}}
 \\ [-0.75mm]
 {\includegraphics[width=.1\linewidth]{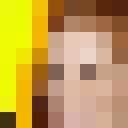}}
& {\includegraphics[width=.1\linewidth]{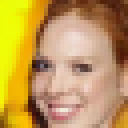}}
& {\includegraphics[width=.1\linewidth]{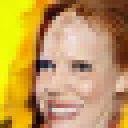}}
& {\includegraphics[width=.1\linewidth]{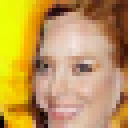}}
& {\includegraphics[width=.1\linewidth]{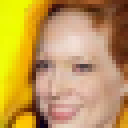}}
& {\includegraphics[width=.1\linewidth]{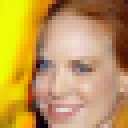}}
& {\includegraphics[width=.1\linewidth]{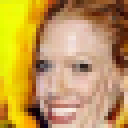}}
& {\includegraphics[width=.1\linewidth]{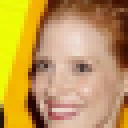}}
 \\ [-0.75mm]
 {\includegraphics[width=.1\linewidth]{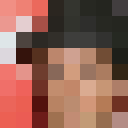}}
& {\includegraphics[width=.1\linewidth]{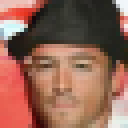}}
& {\includegraphics[width=.1\linewidth]{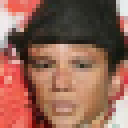}}
& {\includegraphics[width=.1\linewidth]{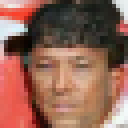}}
& {\includegraphics[width=.1\linewidth]{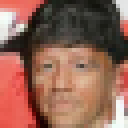}}
& {\includegraphics[width=.1\linewidth]{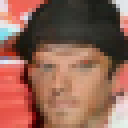}}
& {\includegraphics[width=.1\linewidth]{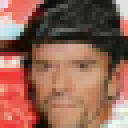}}
& {\includegraphics[width=.1\linewidth]{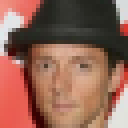}}
 \\ [-0.75mm]
 {\includegraphics[width=.1\linewidth]{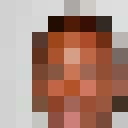}}
& {\includegraphics[width=.1\linewidth]{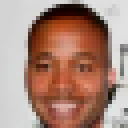}}
& {\includegraphics[width=.1\linewidth]{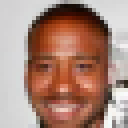}}
& {\includegraphics[width=.1\linewidth]{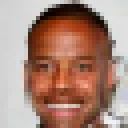}}
& {\includegraphics[width=.1\linewidth]{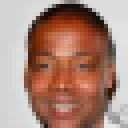}}
& {\includegraphics[width=.1\linewidth]{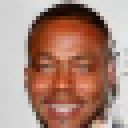}}
& {\includegraphics[width=.1\linewidth]{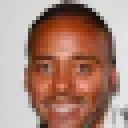}}
& {\includegraphics[width=.1\linewidth]{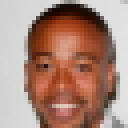}}
 \\ [-0.75mm]
 {\includegraphics[width=.1\linewidth]{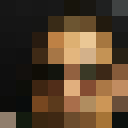}}
& {\includegraphics[width=.1\linewidth]{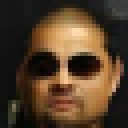}}
& {\includegraphics[width=.1\linewidth]{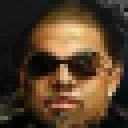}}
& {\includegraphics[width=.1\linewidth]{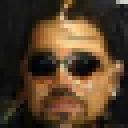}}
& {\includegraphics[width=.1\linewidth]{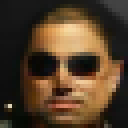}}
& {\includegraphics[width=.1\linewidth]{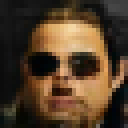}}
& {\includegraphics[width=.1\linewidth]{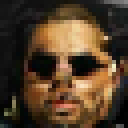}}
& {\includegraphics[width=.1\linewidth]{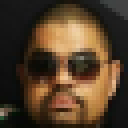}}
\end{tabular} 
\caption{Images from our 1D and 2D local attention super-resolution models trained on CelebA, sampled with different temperatures. 2D local attention with $\tau=0.9$ scored highest in our human evaluation study.}
\end{table*}

\paragraph{CelebA}

We trained both our 1D Local and 2D Local models on the standard CelebA data set of celebrity faces with cropped boundaries. With the 1D Local, we achieve a negative log likelihood (NLL) of $\mathbf{2.68}$ bits/dim on the dev set, using $l_q=128$, memory size of $256$, $12$ self-attention and feed-forward layers, $\modeldim=512$, $8$ attention heads, $2048$ dimensions in the feed-forward layers, and a dropout of $0.1$. With the 2D Local model, we only change the query and memory to now represent a block of size $8\times32$ pixels and $16\times64$ pixels respectively. This model achieves a NLL of $\mathbf{2.61}$ bits/dim.
Existing automated metrics like pSNR, SSIM and MS-SSIM have been shown to not correlate with perceptual image quality \citep{PixelRecursiveSuperResolution}. Hence, we conducted a human evaluation study on Amazon Mechanical Turk where each worker is required to make a binary choice when shown one generated and one real image. Following the same procedure for the evaluation study as \cite{PixelRecursiveSuperResolution}, we show $50$ pairs of images, selected randomly from the validation set, to $50$ workers each. Each generated and original image is upscaled to $128\times128$ pixels using the Bilinear interpolation method. Each worker then has $1$-$2$ seconds to make a choice between these two images. In our method, workers choose images from our model up to $36.1$\% of the time, a significant improvement over previous models. Sampling temperature of $0.8$ and 2D local attention maximized perceptual quality as measured by this evaluation.

To measure how well the high resolution samples correspond to the low resolution input, we calculate Consistency, the $L2$ distance between the low resolution input and a bicubic downsampled version of the high resolution sample. We observe a Consistency score of $0.01$ which is on par with the models in \cite{PixelRecursiveSuperResolution}. 

We quantify that our models are more effective than exemplar based Super Resolution techniques like Nearest Neighbors, which perform a naive look-up of the training data to find the high resolution output. We take a bicubic down-sampled version of our high resolution sample, find the nearest low resolution input image in the training data for that sample, and calculate the MS-SSIM score between the high resolution sample and the corresponding high resolution image in the training data. On average, we get a MS-SSIM score of $44.3$, on $50$ samples from the validation set, which shows that our models don't merely learn to copy training images but generate high-quality images by adding synthesized details on the low resolution input image.

\paragraph{CIFAR-10} We also trained a super-resolution model on the CIFAR-10 data set. Our model reached a negative log-likelihood of $2.76$ using 1D local attention and $2.78$ using 2D local attention on the test set. As seen in Figure~\ref{tab:completion_and_superres}, our model commonly generates plausible looking objects even though the input images seem to barely show any discernible structure beyond coarse shapes.


\section{Conclusion}

In this work we demonstrate that models based on self-attention can operate effectively on modalities other than text, and through local self-attention scale to significantly larger structures than sentences. With fewer layers, its larger receptive fields allow the Image Transformer to significantly improve over the state of the art in unconditional, probabilistic image modeling of comparatively complex images from ImageNet as well as super-resolution.

We further hope to have provided additional evidence that even in the light of generative adversarial networks, likelihood-based models of images is very much a promising area for further research - as is using network architectures such as the Image Transformer in GANs.

In future work we would like to explore a broader variety of conditioning information including free-form text, as previously proposed \citep{Mansimov15}, and tasks combining modalities such as language-driven editing of images.

Fundamentally, we aim to move beyond still images to video \citep{Kalchbrenner16} and towards applications in model-based reinforcement learning.





\bibliography{deeplearn}
\bibliographystyle{icml2018}

\end{document}